\documentclass{article}

\usepackage{arxiv}

\usepackage[english]{babel}  

\usepackage[colorinlistoftodos,prependcaption,textsize=tiny]{todonotes}
\usepackage{graphicx}
\usepackage{latexsym}
\usepackage[nolist]{acronym}
\usepackage{amsmath,amssymb,booktabs}
\usepackage{hyperref}
\usepackage[font=small]{caption}
\usepackage{diagbox}        
\usepackage{adjustbox}
\usepackage[autoload=false,charsperline=100,linenumbers=true,theme=default-plain]{jlcode} 
\usepackage[section]{placeins}
\usepackage{tikz}
\usepackage[numbers]{natbib}  
\usepackage{doi}

\usetikzlibrary{positioning, shapes.geometric, arrows, calc, fit}

\newcommand{\norm}[1]{\left\lVert#1\right\rVert}
\renewcommand{\vec}{\mathbf}

\begin{acronym}
  \acro{ann}[ANN]{artificial neural network}
  \acroplural{ann}[ANNs]{artificial neural networks}
  \acro{mse}[MSE]{mean squared error}
  \acro{nls}[NLS]{non-linear equation system}
  \acro{penode}[PeNODE]{Physics-Enhanced Neural ODE}
  \acroplural{penode}[PeNODEs]{Physics-Enhanced Neural ODEs}
  \acro{pinn}[PINN]{Physics-Informed Neural Network}
  \acro{pca}[PCA]{principal component analysis}
\end{acronym}

\title{Residual-informed Learning of Solutions to Algebraic Loops \thanks{submitted to IDaS-Schriftenreihe, Bielefeld University of Applied Sciences and Arts}}

\date{Oktober 10, 2025}

\usepackage{authblk}

\newbox{\orcid}\sbox{\orcid}{\includegraphics[scale=0.06]{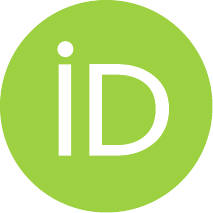}} 
\author[1]{{Felix Brandt}}
\author[2]{
    \href{https://orcid.org/0009-0000-1792-1701}{\usebox{\orcid}\hspace{1mm}Andreas Heuermann}}
\author[1]{
	\href{https://orcid.org/0009-0003-8902-9079}{\usebox{\orcid}\hspace{1mm}Philip Hannebohm}}
\author[1]{
	\href{https://orcid.org/0000-0002-4339-0438}{\usebox{\orcid}\hspace{1mm}Bernhard Bachmann}\thanks{corresponding author}}
\affil[1]{Institute for Data Science Solutions, Bielefeld University of Applied Sciences and Arts, Germany}

\affil[ ]{\texttt{\{felix.brandt,philip.hannebohm,bernhard.bachmann\}@hsbi.de}}

\begin{document}

\maketitle

\begin{abstract}
This paper presents a residual-informed machine learning approach for replacing algebraic loops in equation-based Modelica models with neural network surrogates. A feedforward neural network is trained using the residual (error) of the algebraic loop directly in its loss function, eliminating the need for a supervised dataset. This training strategy also resolves the issue of ambiguous solutions, allowing the surrogate to converge to a consistent solution rather than averaging multiple valid ones. Applied to the large-scale IEEE 14-Bus system, our method achieves a $\sim$60\% reduction in simulation time compared to conventional simulations, while maintaining the same level of accuracy through error control mechanisms.
\end{abstract}


\section{Introduction}
\label{sec:introduction}

Simulating Modelica models can be computationally intensive in the presence of \acp{nls} which typically require iterative methods, such as the Newton-Raphson algorithm, to solve.
As these systems grow larger, the computational cost of solving them becomes significant.

While \ac{ann} surrogates offer fast prediction speed, providing and processing enough high quality labeled training data can be an issue and prohibit high accuracy prediction results.
Reasons can be the size of the dataset, lack of computational resources and/or ambiguities in the provided dataset.
The latter is a big problem in our application.

The residual formulation of an \ac{nls} quantifies the deviation between a proposed solution and a true solution.
This information can be used in the loss function of an \ac{ann} in order to measure goodness of fit, while not requiring a labeled dataset at all.
When the residual is used in the loss function, the \ac{ann} learns to predict the solution vector from an initial guess vector---i.e., a set of approximate values for the algebraic iteration variables---for the algebraic loop.

Still, training an accurate \ac{ann} requires computational resources.
Building such surrogate models is thus particularly useful when a Modelica model needs to be simulated many times, such as in optimization or control tasks.

\subsubsection*{Related Work}

A working automated pipeline for creating surrogate models was presented in~\cite{HEUERMANN2023275}. In that approach, \acp{nls} are replaced with \acp{ann} trained on labeled datasets consisting of input–output pairs. Additionally, they introduced an error control mechanism that falls back to Newton-Raphson when the \ac{ann} prediction error exceeds a predefined threshold. In contrast, this work removes the need for a labeled dataset entirely.

Training a \ac{pinn}~\cite{RAISSI2019686} involves embedding the residual of a partial differential equation (PDE) into the loss function. Our approach is conceptually similar, but instead of using PDE residuals, we use the residuals of algebraic equations that appear in the right-hand side (RHS) computation of ordinary differential equations (ODEs).

As defined in~\cite{dlr200100}, a \ac{penode} represents a meaningful combination of physical equations and \acp{ann}. Our work can be viewed as a special case of \acp{penode}, where the focus lies specifically on replacing algebraic equations with \acp{ann} in a physically consistent manner.

\section{Problem Statement}
\label{sec:problem}

In a Modelica simulation context an algebraic loop is given in residual form
\begin{equation}
    \label{eq:fres}
    \begin{gathered}
        f: \mathbb{R}^{n_{in}} \times \mathbb{R}^{n_{out}} \to \mathbb{R}^{n_{out}},\\
        f(\vec{x},\vec{y}) = \vec{0},
    \end{gathered}
\end{equation}
where $\vec{x} \in \mathbb{R}^{n_{in}}$ contains the simulation time, constant system parameters and other used variables computed in preceding model equations, and $\vec{y} \in \mathbb{R}^{n_{out}}$ are unknowns for which the algebraic loop is solved.
The algebraic loop is solved repeatedly over the course of a simulation with different $\vec{x}$.
To solve \autoref{eq:fres}, an iterative algorithm like Newton's Method is used, which results in
\begin{equation}
    \label{eq:newton}
    \vec{y}_{k+1} := \vec{y}_k - J_f(\vec{x},\vec{y}_k)^{-1} f(\vec{x},\vec{y}_k),
\end{equation}
where $k$ is the current iteration index and $J_f(\vec{x},\vec{y})^{-1}$ is the inverse of the Jacobian of $f$ w.r.t.\ $\vec{y}$.
While Newton's method is effective, it requires repeated computation of the Jacobian and residuals,
which becomes increasingly costly for large systems or systems with complex algebraic loops.

The primary challenge lies in the computational overhead associated with solving non-linear algebraic loops iteratively. This can significantly slow down simulations, especially when the system must be simulated repeatedly for tasks such as optimization or control, thereby motivating the need for more efficient surrogate models.

\begin{figure}[ht]
    \centering
    \begin{tikzpicture}

        \tikzstyle{block} = [rectangle, draw, inner sep=3mm]
        \tikzstyle{arrow} = [thick, ->, >=stealth]

        \node[block] (comp1) at (0, 4) {Preceding Equations};
        \node[block] (nls) at (0.5, 2) {NLS/ANN};
        \node[block] (comp2) at (0, 0) {Following Equations};
        \node[above=1cm of comp1] (input) {states};
        \node[below=1cm of comp2] (output) {derivatives};

        \path[arrow] (input) edge (comp1)
                     ($(comp1.south) + (-1.25, 0)$) edge ($(comp2.north) + (-1.25, 0)$)
                     ($(comp1.south) + (0.5, 0)$) edge node[right] {$\vec{x}$} (nls)
                     (nls) edge node[right] {$\vec{y}$} ($(comp2.north) + (0.5, 0)$)
                     (comp2) edge (output);

        \node[draw, rounded corners, inner sep=5mm, label={65:RHS}, fit=(comp1) (nls) (comp2)] {};

    \end{tikzpicture}
    \caption{The big box represents the computation of the RHS which takes states as inputs and computes the corresponding state derivatives.
    It contains computation of preceding and following equations and specifically one \ac{nls}, that will be replaced by an \ac{ann}.
    The \ac{nls} receives $\vec{x}$ from preceding computations inside the RHS and returns $\vec{y}$.}
    \label{fig:replacement}
\end{figure}
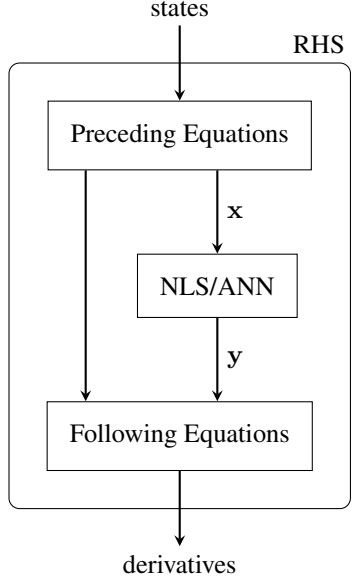

\autoref{fig:replacement} illustrates how the \ac{ann} replaces the \ac{nls} inside the Modelica model.

\subsection{Ambiguity Problem}
\label{sec:ambiguityproblem}

Non-linear algebraic loops can exhibit multiple (ambiguous) solutions, meaning that for a given $\vec{x}$, there may be multiple vectors $\vec{y}$ satisfying $f(\vec{x}, \vec{y}) = 0$. In such cases, $f$ does not define a functional relationship from $\vec{x}$ to $\vec{y}$, as it fails to produce a unique output for each input.

This ambiguity poses a challenge when training neural networks using a standard \ac{mse} loss, since neural networks are designed to approximate continuous functions that assume a unique mapping from inputs to outputs.

The loss formulation proposed in this work addresses this issue by allowing the network to converge to one of the valid solutions, rather than averaging over multiple possibilities. In contrast, training with an \ac{mse} loss in the presence of multiple solutions leads to averaging effects, resulting in poor predictive accuracy. A more detailed discussion and a formal argument can be found in \autoref{sec:ambiguitydiscussion}.

\section{Method}
\label{sec:method}

The approach proposed in this paper uses an \ac{ann} to predict $\vec{y}$ from the input $\vec{x}$, aiming to replace the use of Newton's method for solving algebraic loops. The loss function is defined directly in terms of the residual $f$ as
\begin{equation}
    \label{eq:loss}
    L(\hat{\vec{y}}) = \frac{1}{2} \left\|f(\vec{x}, \hat{\vec{y}})\right\|_2^2.
\end{equation}
Here, $\hat{\vec{y}}$ denotes the predicted solution vector produced by the \ac{ann}, while the true solution vector $\vec{y}$ is unknown and not required for training.

This particular form of $L$ is chosen because it leads to a simpler expression for the gradient, facilitating more efficient training. By minimizing $L$, the \ac{ann} learns to predict $\vec{y}$ such that the residual $f(\vec{x}, \hat{\vec{y}})$ approaches zero.

The specific solution to which the network converges when using \autoref{eq:loss} is influenced by factors such as the random initialization of network weights, the learning rate, and other training parameters—analogous to how Newton's method converges to a particular solution depending on the initial guess vector.

For gradient based optimization such as \ac{ann} training, one needs to compute the gradient of $L$ w.r.t.\ $\hat{\vec{y}}$.
We use the Julia programming language, where automatic differentiation systems like Zygote.jl \cite{Zygote.jl-2018} can handle pure Julia code efficiently.
However, these systems fail when computations involve calls to external functions.
In such cases, the gradient must be derived and implemented manually.
In this study, $f$ is given as C function called from Julia, necessitating the manual implementation of the gradient of $L$.
The gradient of $L$ is
\begin{equation}
    \label{eq:lossgrad}
    \nabla L(\hat{\vec{y}}) = J_f^T(\vec{x},\hat{\vec{y}})f(\vec{x},\hat{\vec{y}}),
\end{equation}
with $J_f^T \in \mathbb{R}^{n_{out} \times n_{out}}$ being the transpose of the Jacobian of $f$ w.r.t.\ $\hat{\vec{y}}$.
$J_f$ is also obtained by a C function call.
$\nabla L(\hat{\vec{y}})$ is of shape $\mathbb{R}^{n_{out}}$.
A detailed derivation of \autoref{eq:lossgrad} is provided in \autoref{sec:lossgrad}. From this point on, we refer to an \ac{ann} trained using the loss function in \autoref{eq:loss} as a \emph{residual-trained model} to emphasize that it learns via residual minimization rather than supervised labels.

An output $\hat{\vec{y}}$ from the \ac{ann} is considered an accurate prediction if the residual $f(\vec{x}, \hat{\vec{y}})$ is sufficiently small, which corresponds to the loss being close to zero. There are two common ways to quantify whether a prediction is “close enough” to a solution.

The first is to use an absolute threshold on the loss:
\[
L(\hat{\vec{y}}) \le \mathit{atol},
\]
where $\mathit{atol}$ is a user-defined absolute tolerance.

The second approach compares successive predictions over the course of training. Let $\hat{\vec{y}}_1$ and $\hat{\vec{y}}_2$ be predictions for a fixed test input after two successive training epochs (or separated by several epochs). Convergence is then defined elementwise by:
\[
|\hat{\vec{y}}_2 - \hat{\vec{y}}_1| \le \mathit{atol} + \mathit{rtol} \cdot |\hat{\vec{y}}_1|,
\]
where $\mathit{rtol}$ is a user-defined relative tolerance that controls convergence independent of the scale of $\hat{\vec{y}}$.
This condition must be satisfied for all components to declare convergence.
Both criteria are implemented and used during training to monitor prediction accuracy.

In addition to residual-based criteria, we also explored using the number of Newton iterations required to reach convergence as an alternative metric. Since our ultimate goal is to reduce the need for such iterations, this provides a meaningful proxy for prediction quality. This criterion could also be used to guide early stopping, e.g., by halting training once the predicted output reduces the number of required Newton steps below a target threshold.

Empirical results supporting this idea are presented in \autoref{sec:experiments}.

One limitation of the current approach is that residuals are not scaled across dimensions. This can lead to imbalance during training when some residual components dominate the loss due to their larger magnitude. Addressing this issue remains part of future work.

For computational efficiency, $\vec{x}$ and $\hat{\vec{y}}$ are typically batched into
matrices with dimensions $\mathbb{R}^{n_{in} \times n}$ and $\mathbb{R}^{n_{out} \times n}$,
respectively.\footnote{Following the Julia convention with the batch size in the last dimension.}
Here $n$ denotes the batch size, or the number of samples.
In this case the loss function becomes
\begin{equation}
    \label{eq:lossbatch}
    L(\hat{\vec{y}}) = \frac{1}{2n} \sum_{i=1}^n \norm{f(\vec{x}[i], \hat{\vec{y}}[i])}_2^2.
\end{equation}
\autoref{eq:lossbatch} is simply the average of \autoref{eq:loss} over the batch.
In the batched case, the gradient is computed by iterating over the batch, evaluating
\autoref{eq:lossgrad} for each element vector, and storing these intermediate results in a
final gradient matrix of shape $\mathbb{R}^{n_{out} \times n}$.
Example implementations are provided in \autoref{sec:implementation}.
An implementation can be found on
GitHub\footnote{GitHub: \href{https://github.com/AMIT-HSBI/UnsupervisedTrainingExample}{AMIT-HSBI/UnsupervisedTrainingExample}}.

\subsection{Semi-Supervised Loss}
\label{sec:semisupervisedloss}

Our method can be extended by adding a supervised \ac{mse} term to \autoref{eq:loss} like so:
\begin{equation}
    \label{eq:semisupervisedloss}
    L(\hat{\vec{y}}, \vec{y}) =
    \frac{\lambda}{2} \norm{f(\vec{x},\hat{\vec{y}})}_2^2
    + \frac{1-\lambda}{2} \norm{\hat{\vec{y}} - \vec{y}}_2^2
\end{equation}
where $\vec{y}$ is the target/label vector for $\vec{x}$.
The hyperparameter $0 \le \lambda \le 1$ controls the relative influence of the two loss terms.
Flux/Zygote handles the automatic differentiation of this modified loss function natively, once the backpropagation rule for the gradient of the first term is provided, which is given in \autoref{eq:lossgrad}.

This approach can be used to guide the learning process towards a specific solution of $f$.
$\vec{y}$ is generated before training with Newton's Method.
A well chosen set of target vectors $\vec{y}$ can be obtained by clustering the label vectors of a randomly generated dataset.
A standard clustering algorithm like k-Means can be used for the task.

\begin{figure}[ht]
    \centering
    \includegraphics[scale=0.4]{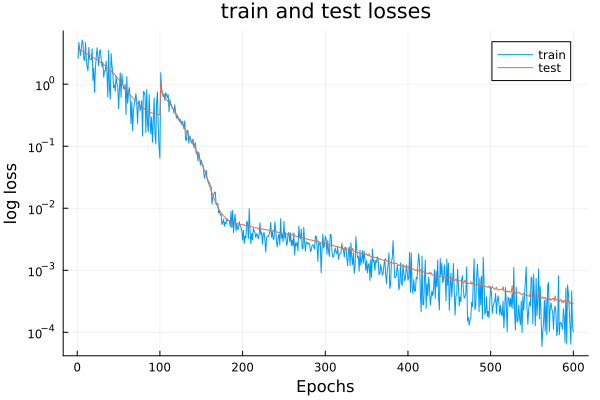}
    \caption{Training loss from a semi-supervised run illustrating the two-phase strategy: MSE loss is used for the first 100 epochs, followed by a residual-based loss. The example is based on the SimpleLoop model from \autoref{sec:simpleloop}.
}
    \label{fig:trainLosssemsup}
\end{figure}

It is also possible to start training using a small unambiguous dataset (obtained through clustering) and
use only the \ac{mse} ($\lambda = 0$) to guide the model towards a preferred direction
and then switch the loss function to the residual ($\lambda \to 1$) to allow training on
a high amount of input data but at the same time guide the training process, which is not possible when only training with \autoref{eq:loss}.
When training like that, we were able to guide the model to a preferred solution in the first phase and in the second phase the model stayed on that solution and reduced the loss further, as seen in \autoref{fig:trainLosssemsup}.







\section{Connected Solutions}
\label{sec:connectedsolutions}

In some cases, the solution set of the \ac{nls} is path-connected, meaning that solution branches are not clearly separated but instead form a continuous manifold. This can lead to the simulation transitioning between solutions over time. However, a single \ac{ann} can typically represent only one solution branch. To address this, we implemented a proof of concept in which multiple \acp{ann} are trained on different regions of the output space, identified using a clustering algorithm such as k-Means. During simulation, the appropriate \ac{ann} is selected based on the input to ensure consistent predictions.



\begin{figure}[ht]
    \centering
    \includegraphics[scale=0.4]{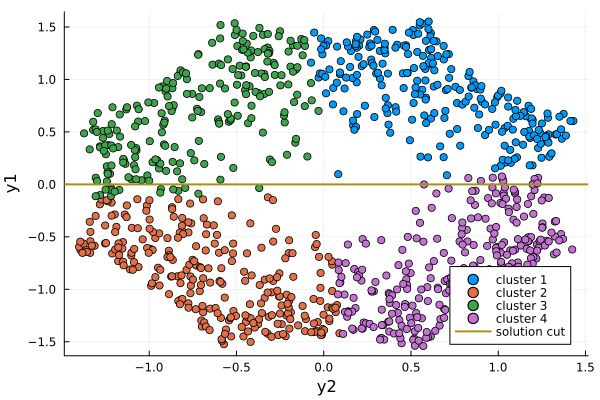}
    \caption{Continuous output space of a model which computes the square root of a complex number, partitioned by k-Means.
    Now a model is trained on each partition and during the simulation one could then only evaluate the model for the part of output space the simulation is currently in.
    The model contains two solutions, and the border between them is indicated by the horizontal ``solution cut'' line.}
    \label{fig:contoutspacegrafikparts}
\end{figure}

The appropriate model is selected as the one trained on the region whose centroid is closest to the previous output. An example implementation of this algorithm, along with resulting figures, is provided in \autoref{sec:appcontoutspace}. 
Using this approach, we demonstrate that employing multiple models significantly improves the system's capability to handle ambiguous problems with connected solution spaces.

\begin{figure}[ht]
    \centering
    \includegraphics[scale=0.4]{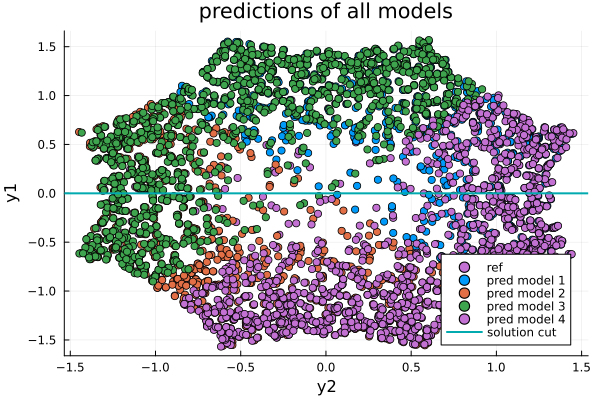}
    \caption{Predictions of all models, each trained on one part of the partition using the second technique from \autoref{sec:semisupervisedloss}.
    One can see that the whole space is covered (with some overlap on the boundaries) and can in principle be predicted.}
    \label{fig:contoutspacegrafikpartspred}
\end{figure}

\section{Experiments and Results}
\label{sec:experiments}

Before training, input data of shape $\mathbb{R}^{n_{in} \times N}$ (with $N \approx 10{,}000$ samples) is generated by sampling the input space. To ensure the samples cover the relevant range, a profiling step first simulates the model once, identifying minimum and maximum bounds for each input variable. This guarantees that the training data spans the full input domain encountered during typical simulations. For more details on this process, see \cite{HEUERMANN2023275}, though as discussed later, this method may be improved.

Sampling is performed using the \texttt{QuasiMonteCarlo.jl} package \cite{QuasiMonteCarlo.jl-2020}, which offers advanced sampling methods designed to provide better coverage of high-dimensional spaces than purely random sampling. We used Sobol sampling, a low-discrepancy sequence that ensures more uniform coverage, reducing the number of samples needed for effective training compared to random sampling. Alternative methods like Latin Hypercube sampling are also available and may be suitable depending on the problem.

To assess the model's generalization performance, validation data is generated by simulating the model once before training, providing unseen examples for evaluation during the training process.

The \ac{ann} architecture used is a feedforward network with two hidden layers of 160 neurons each, employing ReLU activation functions. This configuration balances expressiveness and computational efficiency: enough parameters to capture complex input-output relationships while maintaining fast inference times. Training is performed using the Adam optimizer over up to 2000 epochs, starting with a learning rate of $8 \times 10^{-4}$.

To improve convergence and avoid stagnation, learning rate decay is applied after half the training time. Specifically, the learning rate is multiplied by $\frac{1}{5}$ every 20\% of the remaining epochs. For example, in a 2000-epoch training, the decay steps occur at epochs 1000, 1400, and 1800. 
This gradual reduction helps the model refine its weights during later training stages. While these choices proved effective, there remains significant room for tuning and optimization depending on the specific application.

All training is implemented using the \texttt{Flux.jl} package \cite{Flux.jl-2018} and was run on a CPU (Intel i7-12700T, 1.40 GHz).

The proposed approach is evaluated on two Modelica models that include non-linear algebraic loops. For each model, we assess multiple aspects: the evolution of the training loss, the prediction accuracy during simulation, and the overall simulation time. These results are compared against two baselines: a conventional simulation that solves the algebraic loop using Newton's method, and a supervised surrogate model trained on labeled input-output data. Furthermore, we examine how the approach handles ambiguous solutions, and compare the time required for data generation. In contrast to the supervised method—which requires solving the algebraic loop for each sampled input to generate labels—our method only samples input points, significantly reducing preprocessing cost.

\subsection{SimpleLoop}
\label{sec:simpleloop}

The \textit{SimpleLoop} model describes a growing circle and a moving line, and involves solving a non-linear system to compute the intersection points of the two as they evolve over time. The problem is governed by the following equations:

\begin{equation}
    \label{eq:sldef}
    \begin{gathered}
        x^2 + y^2 = r^2 \\
        x + y = rs
    \end{gathered}
\end{equation}

Here, the first equation defines a circle with radius $r$, and the second defines a line whose position depends on the parameter $s$.

\begin{figure}[ht]
    \centering
    \includegraphics[scale=0.5]{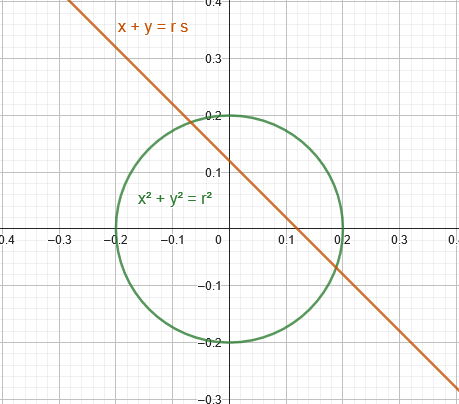}
    \caption{Solution space of \autoref{eq:sldef}. As $r$ varies, the circle grows or shrinks and the line moves along.
            As $s$ varies, the line moves along the main diagonal.}
    \label{fig:slgrafik}
\end{figure}

\begin{figure}[ht]
    \centering
    \begin{tikzpicture}
        \tikzstyle{block} = [draw, rounded corners, inner sep=2mm, minimum height=1cm]

        \node[block] (A) at (0, 0) {$x + y = rs$};
        \node[block] (B) at (4, 0) {$x^2 + y^2 = r^2$};
        \node[block] (Input1) at (0, 2.5) {$r = 1 + t$};
        \node[block] (Input2) at (4, 2.5) {$s=\sqrt{0.9(2-t)}$};

        \node[draw, dashed, rounded corners, inner sep=2mm, label={163:NLS}, fit=(A) (B)] {};

        \path[->] (Input1) edge (A) edge (B)
                  (Input2) edge (A) edge (B)
                  (A) edge[bend left=10] (B)
                  (B) edge[bend left=10] (A);

    \end{tikzpicture}
    \caption{Dependency graph of the SimpleLoop model.
      It shows the circular dependency of the algebraic loop.
      The bottom nodes need to be solved simultaneously for $x$ and $y$.
      Both $r$ and $s$ are known from the preceding equations.}
\end{figure}
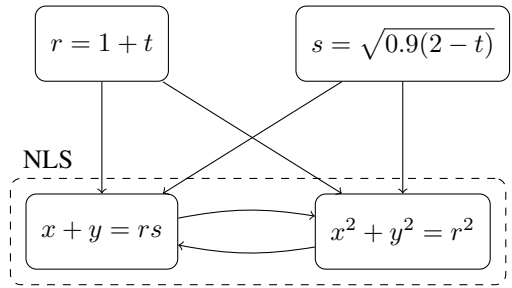

The neural network receives $r$ and $s$ as inputs and is trained to predict the $y$-coordinate of one of the intersection points.
The model is simulated over the interval $t \in [0, 2]$, with $r$ and $s$ varying over time as follows: $r = 1 + t$ and $s = \sqrt{0.9(2 - t)}$.

For this example, the residual function \( f \), introduced in \autoref{eq:fres}, evaluates to
\begin{equation}
    f(\vec{x}, \vec{y}) = y^2 + (rs - y)^2 - r^2,
\end{equation}
where \( \vec{x} = (t, r, s) \), \( \vec{y} = (y) \), and the substitution \( x = rs - y \) has been applied to eliminate \( x \).

This residual formulation results from applying a tearing algorithm, as implemented in OpenModelica
\footnote{\href{https://openmodelica.org/}{www.openmodelica.org}}.
Tearing systematically substitutes equations to reduce the number of unknowns and equations, improving both robustness and efficiency of the numerical solution \cite{Tearing}.

This example serves as a minimal test case for evaluating the ability of the residual-based neural network to learn and solve non-linear algebraic systems embedded in dynamic configurations.

\begin{figure}[ht]
    \centering
    \includegraphics[scale=0.3]{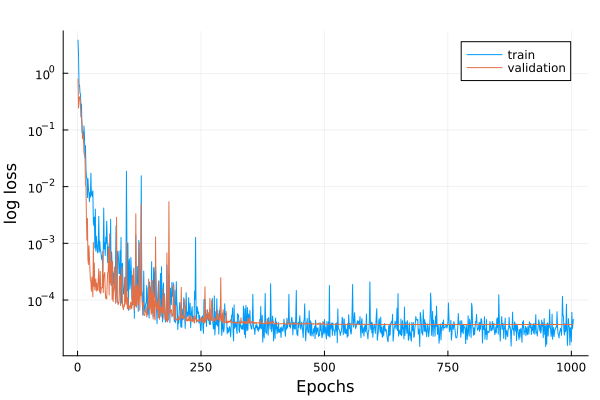}
    \caption{Training and validation loss curve for SimpleLoop.}
    \label{fig:trainLoss}
\end{figure}
\autoref{fig:trainLoss} shows that the model converges early and achieves a training loss of around $1\times10^{-4}$ after 1000 Epochs.
There are no signs of overfitting on the training data.

\begin{figure}[ht]
    \centering
    \includegraphics[scale=0.3]{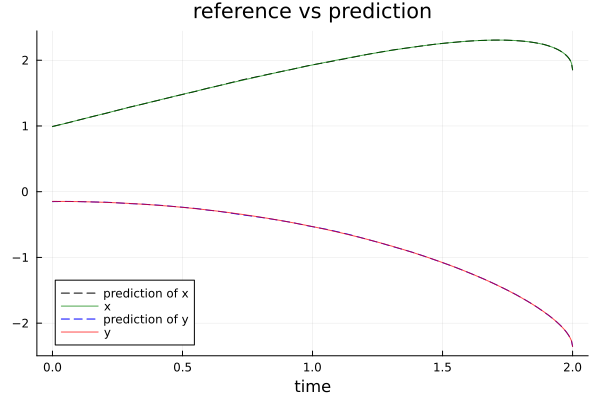}
    \caption{Output variable plot for SimpleLoop. The model outputs a prediction for
        $y$ and one can compute the corresponding prediction for $x$.}
    \label{fig:iterationVarSolution}
\end{figure}
In \autoref{fig:iterationVarSolution}, the predicted and reference trajectories are visually indistinguishable.


\begin{figure}[ht]
    \centering
    \includegraphics[scale=0.35]{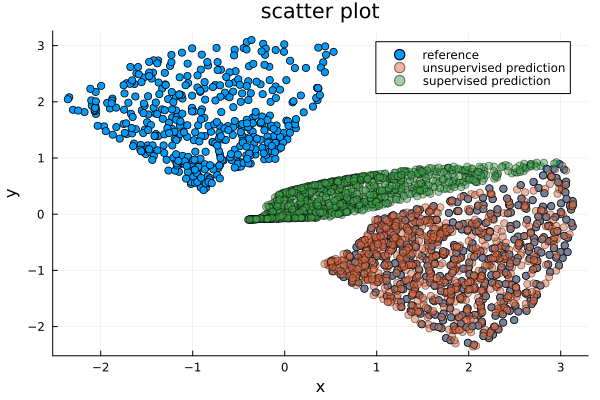}
    \caption{Scatter Residual/Supervised comparison plot for SimpleLoop.}
    \label{fig:scatterSL}
\end{figure}
As shown in \autoref{fig:scatterSL}, the supervised model (green) averages between the two true solutions (blue), making it unsuitable for ambiguous problems. In contrast, the residual-based model consistently converges to one of the valid solutions. A potential improvement to the supervised approach could involve clustering the dataset prior to training and training on one of the solution branches individually.

\begin{table}[ht]
    \caption{
    Simulation time (in seconds) for the SimpleLoop model versus number of training epochs.
    \textit{Residual} refers to our method using a surrogate trained with the residual-based loss (\autoref{eq:loss}).
    \textit{Supervised} uses a surrogate trained on labeled data with an MSE loss.
    \textit{Classical} denotes the standard simulation using Newton's method.
    All surrogates were trained on 10{,}000 data samples.
    }

    \centering
    \begin{tabular}{@{}lrrr@{}}
        \toprule
                    & \multicolumn{3}{c}{Epochs} \\ \cmidrule{2-4}
        Method      & 10      & 100     & 1000 \\
        \midrule
        Residual             & 1.001 s  & 1.001 s  & 1.001 s  \\
        Supervised           & 1.001 s  & 1.001 s  & 1.001 s  \\
        \midrule
        Classical            &          & 1.001 s  &   \\
        \bottomrule
    \end{tabular}
    \label{tab:simTimeSL}
\end{table}

\autoref{tab:simTimeSL} shows no significant difference in simulation time between the three approaches for this example.
To produce the results, three sets of five simulation runs were performed for each number of training epochs and method.
The simulation times were measured using Julia's \texttt{time()} function and averaged over the five runs.

During surrogate-based simulation, the relative error of the predicted solution is monitored.
If the error exceeds a user-defined tolerance, Newton's Method is invoked using the predicted value as the initial guess.
This mechanism ensures that surrogate use does not compromise accuracy.

In this particular case, all approaches achieved similar simulation times, indicating no performance gain from the surrogate due to the simplicity of the model.


\begin{table}[ht]
    \caption{Average data generation time in milliseconds over 5 runs for different numbers of samples $N$.
    \textit{Supervised} refers to generating a labeled dataset (input-output pairs), while \textit{Residual} denotes our method, which only samples input points.}

    \centering
    \begin{tabular}{@{}lrrrr@{}}
        \toprule
                    & \multicolumn{4}{c}{$N$} \\ \cmidrule{2-5}
        Method      & 100       & 1000      & 10000    & 100000 \\
        \midrule
        Supervised  & 8340 ms   & 8410 ms   & 8580 ms  & 9010 ms \\
        Residual    & \bf 0.0176 ms & \bf 0.0494 ms & \bf 0.439 ms & \bf 7.78 ms \\
        \bottomrule
    \end{tabular}
    \label{tab:genTimeSL}
\end{table}

The residual data generation process involves only sampling input points within the previously determined bounds, which is computationally inexpensive. In contrast, the supervised approach introduces additional overhead, as it requires solving the non-linear system using Newton's method for each sampled input point to generate corresponding output labels. This can be observed in \autoref{tab:genTimeSL}. While the supervised method does not scale significantly with the number of samples $N$, its per-sample cost remains higher due to this additional computation.

The proposed residual-based approach accurately captured the behavior of the algebraic loop in the \textit{SimpleLoop} model and yielded satisfactory results in terms of training loss and agreement with reference values. However, as this is a relatively simple, illustrative example, no measurable improvement in simulation time over the classical approach was observed.

\subsection{IEEE14}
\label{sec:iee14}

The \textit{IEEE14} bus system is an approximate model of the American electric power grid as it existed in February 1962~\cite{IEEE14}, and it is included in the OpenIPSL Modelica library~\cite{BAUDETTE201834}. The model contains multiple \acp{nls} that can be targeted for surrogate replacement. In this study, we investigate the substitution of two \acp{nls} of differing sizes to evaluate the performance of our approach on more complex and realistic systems.

\subsubsection{Large system}

This \ac{nls} features a 16-dimensional input vector, referred to as the ``used variables'' $\vec{x}$, and produces a 110-dimensional output vector of ``iteration variables'' $\vec{y}$. To substitute this system, we design a neural network with matching input and output dimensions.

\begin{figure}[ht]
    \centering
    \includegraphics[scale=0.35]{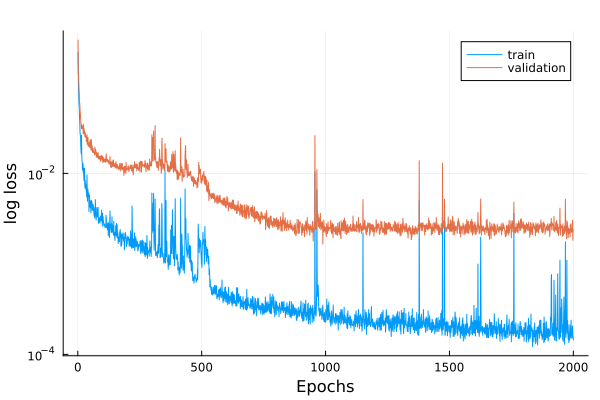}
    \caption{Training and validation loss curve for IEEE14.}
    \label{fig:ieee14loss}
\end{figure}

\autoref{fig:ieee14loss} illustrates that the model steadily converges toward a minimum. However, the plot also reveals signs of overfitting and occasional undesired loss spikes.

\begin{figure}[ht]
    \centering
    \includegraphics[scale=0.4]{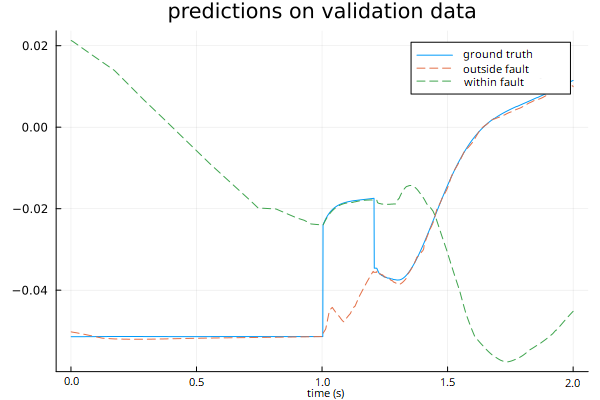}
    \caption{Time series \textit{ground truth} of one reference iteration variable compared to predictions from two neural networks: Model \textit{outside fault} trained only outside the ground fault and model \textit{within fault} trained during the simulated ground fault.}
    \label{fig:itvarieee141403}
\end{figure}
The plot over time in \autoref{fig:itvarieee141403} clearly shows two distinct events corresponding to a ground fault starting at 1 second and lasting for 200 ms. Before the fault, the variable path changes only slightly, appearing almost constant.

The ground fault manifests in the model as the residual depends on an if-statement conditioned on the current simulation time, which introduces another form of ambiguity by switching between branches of the piecewise-defined equation.
To handle this, we train two separate neural networks: one with the ground fault switched off, and another during the fault. Consequently each network models a different branch of the if-statement.

The results, also shown in \autoref{fig:itvarieee141403}, demonstrate that one model accurately predicts behavior during the ground fault, while the other performs well before and after it. The final prediction is constructed by combining the outputs of both networks appropriately.

During actual simulation, the appropriate model would be selected dynamically based on the same condition used in the original residual formulation.



\begin{figure}[ht]
    \centering
    \includegraphics[scale=0.4]{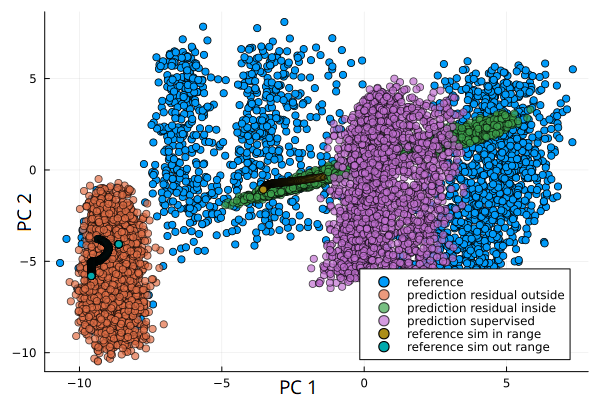}
    \caption{\acs{pca} plot of the large IEEE14 system's output space, comparing reference data with residual and supervised predictions. The residual prediction is separated into outputs from the models trained during and outside the ground fault. The output space dimensionality was reduced to two principal components using \acs{pca}. Also shown are the data points traversed by a reference simulation.}
    \label{fig:pcaIEEE14refpred}
\end{figure}
\autoref{fig:pcaIEEE14refpred} reveals that the simulation data clusters with \ac{pca} into two distinct groups, supporting the use of separate networks to handle the ground fault scenario. The supervised model's predictions, however, do not converge to a single solution; instead, they spread between clusters, indicating ambiguity. In contrast, each unsupervised (residual) model converges clearly to one solution cluster without averaging across clusters. This suggests that our method effectively resolves ambiguity both in low- and high-dimensional spaces.

Additionally, it is evident that many reference data points contribute little to training, highlighting the potential benefit of a more targeted data generation strategy focused on regions near actual simulation trajectories.

\autoref{tab:newtoniterations} shows the progression of the number of Newton iterations required for convergence throughout the training process. The first data point corresponds to the completely untrained network, and subsequent measurements are taken every 50 epochs during training.

\begin{table}[ht]
    \caption{Number of Newton iterations to converge when initialized with the model prediction. The first column shows iterations starting from a random initial guess without model prediction. Subsequent values represent the mean iterations over a batch of inputs, typically between 3 and 4.}

    \centering
    \begin{tabular}{@{}lrrrr@{}}
        \toprule
                    & \multicolumn{4}{c}{Num Epochs} \\ \cmidrule{2-5}
        Model      & 0 & 100 & 500 & 1000  \\
        \midrule
        Large system & 186.7 & 4.0 & 3.7 & \bf{3.3} \\
        \bottomrule
    \end{tabular}
    \label{tab:newtoniterations}
\end{table}

In \autoref{tab:newtoniterations}, a clear improvement is observed over the course of training, with the number of Newton iterations converging to fewer than five. Notably, after an initial drop, the required iterations stabilize, suggesting that stopping training early would not significantly affect the eventual speedup in simulation time. Consequently, whether training lasts for 100 or 1000 epochs may have little impact if the primary goal is to improve simulation efficiency.

\begin{table}[ht]
    \caption{Simulation time in seconds versus number of training epochs for the IEEE14 model. Classical refers to conventional simulation; Surrogate denotes our method. Training used 5000 data samples.}
    \centering
    \begin{tabular}{@{}lrrr@{}}
        \toprule
                    & \multicolumn{3}{c}{Epochs} \\ \cmidrule{2-4}
        Method      & 10      & 100      & 1000   \\
        \midrule
        Residual             & \bf 10.0 s & \bf 10.0 s & \bf 10.2 s \\
        Supervised           & 14.0 s & \bf 10.0 s & \bf 10.2 s \\
        \midrule
        Classical                & & 28.0 s &    \\
        \bottomrule
    \end{tabular}
    \label{tab:simTimeIEEE14}
\end{table}
\autoref{tab:simTimeIEEE14} demonstrates that simulation using the surrogate model is approximately 60\% faster than the conventional approach.

\begin{figure}[ht]
    \centering
    \includegraphics[scale=0.45]{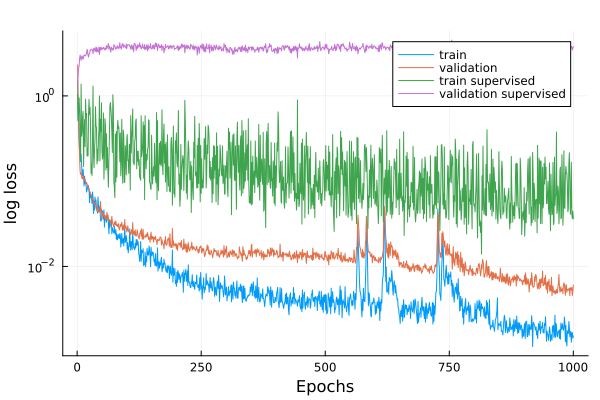}
    \caption{Comparison of training and validation loss progressions for supervised and residual methods. The residual approach clearly outperforms the supervised method, as expected.}
    \label{fig:loss_sup_res_comp_ieee}
\end{figure}

\autoref{fig:loss_sup_res_comp_ieee} compares the training and validation loss curves of the supervised and residual methods. 
The residual method converges to a lower loss than the supervised method. This is expected, as shown in e.g. \autoref{fig:scatterSL}.

\begin{table}[ht]
    \caption{Data generation time (in seconds) for varying sample sizes $N$. Residual denotes our method, while Supervised refers to generating labeled datasets.}
    \centering
    \begin{tabular}{@{}lrrr@{}}
        \toprule
                    & \multicolumn{3}{c}{$N$} \\ \cmidrule{2-4}
        Method      & 100        & 1000       & 10000      \\
        \midrule
        Supervised  & 44.347     & 58.047     & 449.467 \\
        Residual    & \bf 0.0005 & \bf 0.0003 & \bf 0.0003 \\
        \bottomrule
    \end{tabular}
    \label{tab:genTiemIEEE14}
\end{table}
\autoref{tab:genTiemIEEE14} shows that supervised data generation requires significantly more time, as expected. Due to the high dimensionality of the system, issues such as NaN and infinite values during Newton iterations occur more frequently, leading to prolonged generation times.

\subsubsection{Smaller system}
Here, we replace a smaller algebraic loop within the IEEE14 model, characterized by four input variables and a single output.

\begin{figure}[ht]
    \centering
    \includegraphics[scale=0.4]{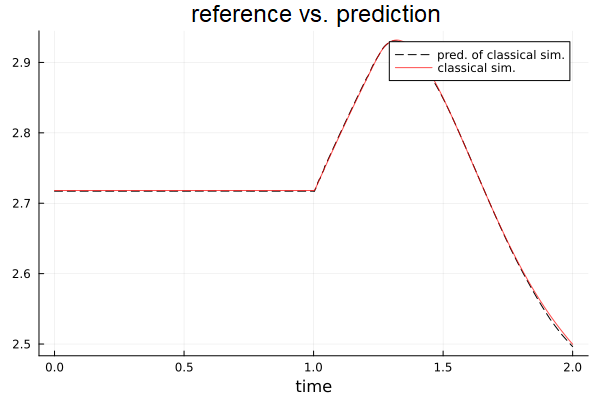}
    \caption{Prediction versus reference plot of the single output dimension over simulation time.}
    \label{fig:itvar1594}
\end{figure}

\autoref{fig:itvar1594} shows high prediction accuracy, with the surrogate model closely matching the reference output throughout the simulation. This indicates that the network effectively captures the system's nonlinear behavior for this smaller algebraic loop, demonstrating the capability of the proposed residual training approach in delivering precise predictions even in low-dimensional cases.

\section{Summary and Outlook}
\label{sec:summary}

This publication explored the application of a residual loss approach to train neural networks for learning algebraic loops. The method proved successful on both small-scale toy problems, such as \textit{SimpleLoop}, and more complex systems like the \textit{IEEE 14-bus} network, achieving accurate predictions and improved simulation performance.

A key strength of this approach lies in its ability to eliminate the need for labeled datasets, thereby significantly accelerating the data preparation phase. Furthermore, it effectively addresses the ambiguity problem often encountered when using supervised loss functions like \ac{mse}, particularly in systems that admit multiple valid solutions.

Several directions for further improvement emerged from this work. One promising enhancement involves adaptive sampling, where input points are actively selected in regions of high error to increase accuracy and better capture complex behaviors in the input space. Another avenue is domain decomposition, which would partition the input domain into subregions and train specialized models for each, potentially reducing model complexity and enhancing generalization.

Additionally, guiding the sampling process with reference trajectories---those closer to actual simulation behavior---may yield more data-efficient training and faster convergence. Addressing overfitting remains an important concern, especially in larger or more ambiguous systems, where generalization is critical to ensure robustness. Lastly, the current training process does not include residual scaling; incorporating appropriate scaling of residual terms, particularly when dealing with equations of differing magnitudes or units, could improve training stability, convergence speed, and overall accuracy.

Future work will focus on these directions to further refine the proposed method and extend its applicability to real-world simulation tasks involving algebraic loops. These advancements hold the potential to improve both simulation fidelity and computational efficiency in complex hybrid modeling environments.


%
%

\newpage
\onecolumn
\appendix

\section{Example Implementation}
\label{sec:implementation}

Example implementation using Julia v1.10.4 and
ChainRulesCore.jl\footnote{https://github.com/JuliaDiff/ChainRulesCore.jl} v1.24.0

\jlinputlisting[language=julia, style=jlcodestyle]{juliaSources/loss.jl}

\section{Gradient of Residual Loss Function}
\label{sec:lossgrad}

Derivation of \autoref{eq:lossgrad}:
\begin{align*}
    L(\hat{\vec{y}})
    = \frac{1}{2} \norm{f(\vec{x},\hat{\vec{y}})}_2^2
    = \frac{1}{2} \left(\sqrt{\sum_{i=1}^n f_i(\vec{x},\hat{\vec{y}})^2}\right)^2
    = \frac{1}{2} \sum_{i=1}^n f_i(\vec{x},\hat{\vec{y}})^2
\end{align*}
Taking the derivative w.r.t.\ $\hat{\vec{y}}$ gives
\begin{align*}
    \nabla L(\hat{\vec{y}})
    &= \nabla \left( \frac{1}{2} \sum_{i=1}^n f_i(\vec{x},\hat{\vec{y}})^2 \right)
    = \frac{1}{2} \sum_{i=1}^n 2 f_i(\vec{x},\hat{\vec{y}}) \nabla f_i(\vec{x},\hat{\vec{y}})
    = \sum_{i=1}^n f_i(\vec{x},\hat{\vec{y}}) \nabla f_i(\vec{x},\hat{\vec{y}}) \\
    &= f_1(\vec{x},\hat{\vec{y}})
    \begin{pmatrix}
        \frac{\partial f_1}{\partial \hat{y}_1} \\
        \vdots \\
        \frac{\partial f_1}{\partial \hat{y}_n}
    \end{pmatrix}
    + \dots + f_n(\vec{x},\hat{\vec{y}})
    \begin{pmatrix}
        \frac{\partial f_n}{\partial \hat{y}_1} \\
        \vdots \\
        \frac{\partial f_n}{\partial \hat{y}_n}
    \end{pmatrix}
    = \begin{pmatrix}
        \frac{\partial f_1}{\partial \hat{y}_1} & \ldots & \frac{\partial f_n}{\partial \hat{y}_1} \\
        \vdots & \ddots & \vdots \\
        \frac{\partial f_1}{\partial \hat{y}_n} & \ldots & \frac{\partial f_n}{\partial \hat{y}_n}
    \end{pmatrix}
    \begin{pmatrix}
        f_1(\vec{x},\hat{\vec{y}}) \\
        \vdots \\
        f_n(\vec{x},\hat{\vec{y}})
    \end{pmatrix} \\
    &= J_f^T(\vec{x},\hat{\vec{y}}) f(\vec{x},\hat{\vec{y}})
\end{align*}

\newpage
\section{Ambiguity discussion}
\label{sec:ambiguitydiscussion}

Consider an ambiguous dataset $D = \{(x, y_i)\}_{i=1}^n$, where
$n$ is the number of data points in $D$, and there exist multiple distinct target values $y_i$ corresponding to the same input $x$, i.e., there exist at least two indices $i \neq j$ such that $y_i \neq y_j$.

We argue that a multilayer perceptron (MLP) $f_{\theta}$ trained on $D$ using a mean squared error (\ac{mse}) loss cannot achieve arbitrary precision. Instead, its prediction for $x$ will converge to the average of all $y_i$, i.e.,
\[
\hat{y} = f_{\theta}(x) = \frac{1}{n} \sum_{i=1}^n y_i.
\]
This implies that the minimal achievable loss is bounded below by a nonzero quantity due to the inherent ambiguity in the dataset.

More formally, since $f_{\theta}$ is a function and the dataset $D$ does not represent a function (as it maps a single input $x$ to multiple outputs $y_i$), it follows that $f_{\theta}$ cannot perfectly fit all target values simultaneously. Thus, the best the model can do under \ac{mse} is to output a central value—specifically, the mean of all $y_i$—which minimizes the squared error.

We now aim to formally show that this averaging behavior indeed arises by minimizing the following objective:
\begin{align*}
    L(\hat{y}, y) = \frac{1}{n}\sum_{i=1}^n(\hat{y}_i - y_i)^2.
\end{align*}
Since all input values $x$ are equal, the corresponding predictions $\hat{y}_i = f_\theta(x) = \hat{y}$ are also equal. We therefore write:
\begin{align*}
    L(\hat{y}, y) = \frac{1}{n}\sum_{i=1}^n(\hat{y} - y_i)^2.
\end{align*}
To find the minimum, we take the gradient with respect to $\hat{y}$ and set it to zero:
\begin{align*}
    \nabla_{\hat{y}} L(\hat{y}, y) = \frac{2}{n}\sum_{i=1}^n(\hat{y} - y_i) = 0
    \quad \Rightarrow \quad \sum_{i=1}^n(\hat{y} - y_i) = 0
    \quad \Rightarrow \quad n\hat{y} = \sum_{i=1}^n y_i
    \quad \Rightarrow \quad \hat{y} = \frac{1}{n}\sum_{i=1}^n y_i.
\end{align*}
Thus, the MSE loss is minimized when the network output $\hat{y}$ equals the mean of all target values.

Now we compute the corresponding minimum value of the loss by inserting $\bar{y} = \frac{1}{n} \sum_{i=1}^n y_i$ into the loss function:
\begin{align*}
    L(\bar{y}, y) &= \frac{1}{n}\sum_{i=1}^n(\bar{y} - y_i)^2
    = \frac{1}{n}\sum_{i=1}^n(\bar{y}^2 - 2\bar{y}y_i + y_i^2)
    = \frac{1}{n} \left( n\bar{y}^2 - 2\bar{y} \sum_{i=1}^n y_i + \sum_{i=1}^n y_i^2 \right)\\
    &= \bar{y}^2 - 2\bar{y}^2 + \frac{1}{n} \sum_{i=1}^n y_i^2
    = \frac{1}{n} \sum_{i=1}^n y_i^2 - \bar{y}^2.
\end{align*}
Therefore, the minimum achievable loss is given by:
\begin{align*}
    L(\bar{y}, y) = \frac{1}{n} \sum_{i=1}^n y_i^2 - \left( \frac{1}{n} \sum_{i=1}^n y_i \right)^2,
\end{align*}
which is the \emph{empirical variance} of the $y_i$ values. Since the dataset is ambiguous (i.e., not all $y_i$ are equal), the variance is strictly positive:
\begin{align*}
    L(\bar{y}, y) > 0.
\end{align*}

\noindent
This proves that a neural network trained using MSE on ambiguous data will output the mean of the targets and cannot reduce the loss arbitrarily—highlighting a key limitation in such settings.

It is now evident that in \autoref{eq:loss}, the loss $L(\hat{\vec{y}})$ satisfies $L(\hat{\vec{y}}) = 0$ if and only if \autoref{eq:fres} is fulfilled. Since the loss function is always non-negative, this condition also corresponds to a global minimum.

An optimization algorithm such as Adam will naturally converge to one of the minima of the loss landscape. In the presence of multiple local minima—arising, for instance, from ambiguous solutions—the optimizer will simply settle on one of them.

Therefore, ambiguities are inherently handled by our approach: instead of averaging across all possible outputs (as in the supervised case), the network trained using the residual loss from \autoref{eq:loss} will converge to a valid solution that satisfies the residual equation, avoiding the undesirable averaging effect.

\pagebreak
\section{Continuous output space}
\label{sec:appcontoutspace}

\jlinputlisting[language=julia, style=jlcodestyle]{juliaSources/constr_sol.jl}

\begin{figure}[ht]
    \centering
    \includegraphics[width=.5\textwidth]{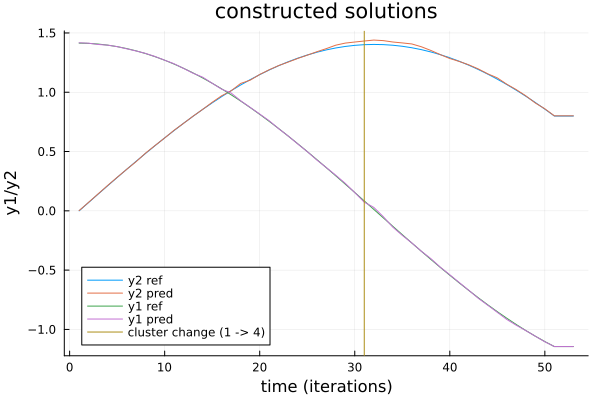}
    \caption{Constructed solutions for both output variables in the complex square root example.
    The vertical line marks the transition from cluster 1 to cluster 4 during simulation; after this point, the model trained on cluster 4 is used for prediction.}
    \label{fig:constructedsolsy1y2fig}
\end{figure}

\begin{figure}[ht]
    \centering
    \includegraphics[width=.5\textwidth]{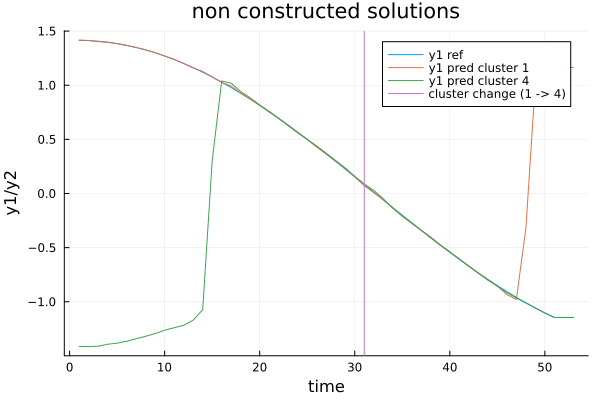}%
    \includegraphics[width=.5\textwidth]{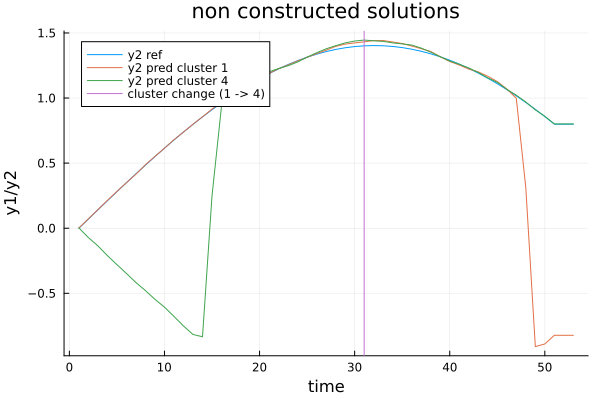}
    \caption{Non-constructed solutions for the output variables $y_1$, $y_2$.
    At the beginning of the simulation, the model trained on cluster 4 does not follow the reference trajectory; toward the end, the model trained on cluster 1 fails to do so.
    This illustrates that using either model alone would be insufficient, highlighting the utility of the switching algorithm.}
    \label{fig:nonconstructsoly2fig}
\end{figure}

%
%

\end{document}